\documentclass{llncs}
\usepackage{amsmath}
\usepackage{graphicx}
\usepackage{epstopdf}
\usepackage{amssymb}
\usepackage{subfigure}
\usepackage{cite}
\usepackage{fancyhdr}
\usepackage{algpseudocode}
\usepackage{algorithm}
\usepackage{multirow}
\usepackage{booktabs}

\begin{document}
\title{Automatic Vertebra Labeling in Large-Scale 3D CT using Deep Image-to-Image Network with Message Passing and Sparsity Regularization}
\newcommand*\samethanks[1][\value{footnote}]{\footnotemark[#1]}
 \author{Dong Yang\inst{1}\thanks{Authors contributed equally}, Tao Xiong\inst{2}\samethanks, Daguang Xu\inst{3}\thanks{\{daguang.xu, shaohua.zhou, dorin.comaniciu\}@siemens-healthineers.com}, Qiangui Huang\inst{4}, David Liu\inst{3}, S. Kevin Zhou\inst{3}\samethanks, Zhoubing Xu\inst{3}, JinHyeong Park\inst{3}, Mingqing Chen\inst{3}, Trac D. Tran\inst{2}, Sang Peter Chin\inst{2}, Dimitris Metaxas\inst{1} \and Dorin Comaniciu\inst{3}\samethanks}
 \institute{Department of Computer Science, Rutgers University, Piscataway, NJ 08854, USA
 \and
 Department of Electrical and Computer Engineering, The Johns Hopkins University, Baltimore, MD 21218, USA
 \and
 Medical Imaging Technologies, Siemens Healthcare Technology Center, Princeton, NJ 08540, USA
 \and
 Department of Computer Science, University of Southern California, LA, California 90089, USA}
\maketitle

\begin{abstract}
Automatic localization and labeling of vertebra in 3D medical images plays an important role in many clinical tasks, including pathological diagnosis, surgical planning and postoperative assessment. However, the unusual conditions of pathological cases, such as the abnormal spine curvature, bright visual imaging artifacts caused by metal implants, and the limited field of view, increase the difficulties of accurate localization. In this paper, we propose an automatic and fast algorithm to localize and label the vertebra centroids in 3D CT volumes. First, we deploy a deep image-to-image network (DI2IN) to initialize vertebra locations, employing the convolutional encoder-decoder architecture together with multi-level feature concatenation and deep supervision. Next, the centroid probability maps from DI2IN are iteratively evolved with the message passing schemes based on the mutual relation of vertebra centroids. Finally, the localization results are refined with sparsity regularization. The proposed method is evaluated on a public dataset of 302 spine CT volumes with various pathologies. Our method outperforms other state-of-the-art methods in terms of localization accuracy. The run time is around 3 seconds on average per case. To further boost the performance, we retrain the DI2IN on additional \emph{1000}+ 3D CT volumes from different patients. To the best of our knowledge, this is the first time more than \emph{1000} 3D CT volumes with expert annotation are adopted in experiments for the anatomic landmark detection tasks. Our experimental results show that training with such a large dataset significantly improves the performance and the overall identification rate, for the first time by our knowledge, reaches \emph{90}\%.

\end{abstract}
\section{Introduction}
\begin{figure}[t]
    \centering
	\vspace{-2mm}
    \centerline{\includegraphics[width=0.7\columnwidth]{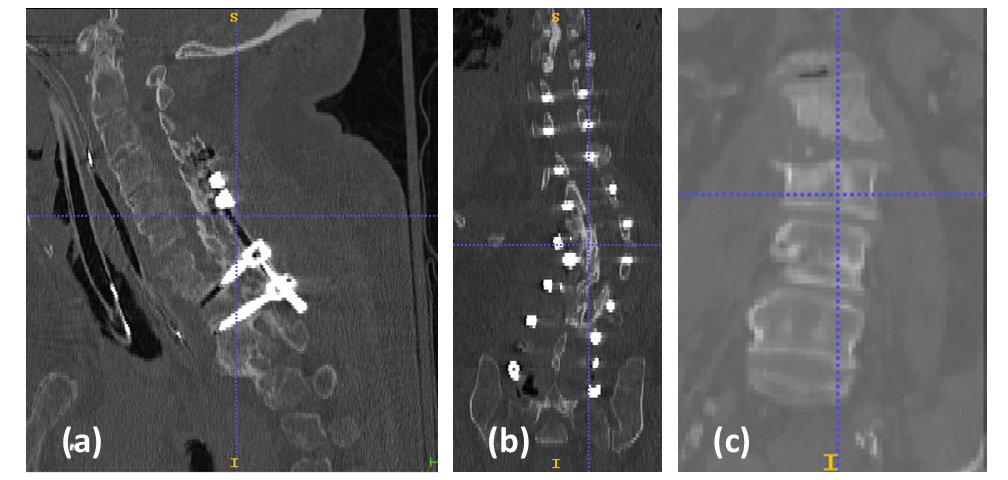}}
	\vspace{-2mm}
    \caption{Demonstration of pathological cases. (a) Surgical Metal Implants (b) Spine Curvature (c) Limited FOV}
	\vspace{-5mm}
    \label{fig:1}
\end{figure}

Automatic localization and labeling of vertebrae in 3D spinal imaging, e.g. computed tomography (CT) or magnetic resonance imaging (MRI), has become an essential tool for clinical tasks, including pathological diagnosis, surgical planning and post-operative assessment. Specific applications such as vertebrae segmentation, fracture detection, tumor detection, registration and statistical shape analysis can also benefit from the effective vertebrae detection and labeling algorithms. However, there are many challenges associated with designing an accurate and automatic algorithm, which arise from pathologies, image artifacts, and the limited field-of-view. For example, as shown in Figure \ref{fig:1}, the abnormal spine curvature and surgical metal implants significantly alter the appearance of vertebrae and reduce the image contrast. Spine-focused scans with small field-of-view (FOV) also add difficulty to the identification tasks due to lack of global spatial and contextual information.

To address these challenges, many approaches have been proposed for automatic localization and identification of vertebrae. Glocker \emph{et al.}\cite{glocker2012automatic} presented a method based on regression forests and probabilistic graphic models. However, their method is likely to suffer from the narrow field-of-view because the broad contextual information is not always available. To overcome this limitation, Glocker \emph{et al.}\cite{glocker2013vertebrae} proposed a randomized classification forest based approach which achieved reasonable localization and identification performances on pathological cases and those with limited FOV. Recently, deep learning has been employed in the applications of spine detection. Chen \emph{et al.}\cite{chen2015automatic} presented a joint convolutional neural network (J-CNN). This hybrid approach used a random forest classifier to coarsely localize the candidates before the J-CNN scaned the input CT volume for final results. By incorporating the pairwise information of neighboring vertebrae in J-CNN, it outperformed other methods\cite{glocker2013vertebrae}. Suzani \emph{et al.}\cite{suzani2015fast} proposed a deep feed-forward neural network to detect if an input image contained a specific vertebra. Although this work achieved high detection rates, it reported a large mean localization error compared with other works. Besides, instead of the direct 3D volumetric input, this work extracted 1D features based on the local voxel intensities as the input of deep feed-forward neural network. In addition, no convolution or pooling operation was applied in the network. Payer \emph{et al.}\cite{payer2016regressing} proposed a composite neural network to build up the full connection between response maps of all landmarks with convolutional kernels. The spatial relationship of landmarks were implicitly embedded in the CNN model.

In order to overcome these limitations and to take advantage of deep neural networks, we present an approach, shown in Figure \ref{fig:pipeline}, with the following contributions:

 \begin{figure}[t]
    \centering
	\vspace{-2mm}
    \centerline{\includegraphics[width=0.8\columnwidth]{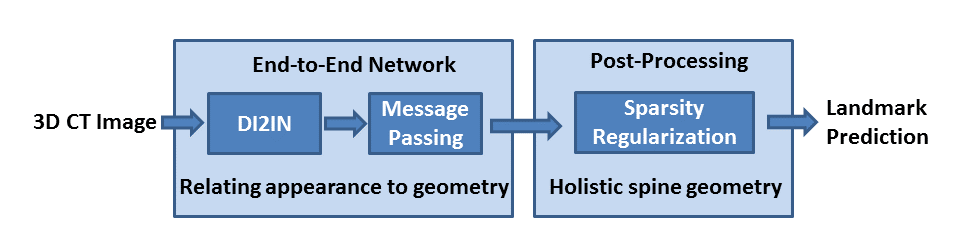}}
	\vspace{-2mm}
    \caption{Proposed method which consists of three major components: deep Image-to-Image Network (DI2IN), message passing and sparsity regularity.}
	\vspace{-5mm}
    \label{fig:pipeline}
\end{figure}

\emph{a) Deep Image-to-Image Network (DI2IN) for Voxel-Wise Regression}

Without extracting features from input images, the proposed deep image-to-image architecture directly takes a 3D CT volume as input. The training of the proposed network is designed as multichannel voxel-wise regression (refer to section 2.1). It generates the multichannel probability maps associated with different vertebra centers, which intuitively illustrate the location and label of vertebrae. Our neural network requires no coarse classifiers to remove the outliers for preprocessing. Instead, it automatically extracts contextual and spatial information by itself. By taking the advantage of fully convolutional implementation, the proposed network is significantly time-efficient, which sets it apart from the sliding window approaches.

\emph{b) Response Enhancement with Message Passing}

Although the proposed deep image-to-image network generates confident probability maps, there is no guarantee that it will avoid false positives (outliers) due to the complexity of appearance (shown in Figure \ref{fig:1}).
To resolve this problem, we adopt a message passing scheme within the probability maps of vertebra centers, which leverages the mutual relation of vertebrae.
A chain-structure graphical model is introduced to depict the spatial relationship.
Each node in the model represents a probability distribution of one vertebra center.
During the passing scheme, the probability map of each vertebra center iteratively receives messages (encoded in the convolution operation) from all neighboring vertebrae (nodes) and absorbs them for further self-evolvement.
The collected messages can not only enhance the response of correct location, but also suppress that of the false positives.

\emph{c) Refinement using Sparse Representation}

To further refine the coordinates of vertebrae, we incorporate a dictionary learning and sparse representation approach which utilizes the holistic structure of the spine and identifies the important set of coordinates. Instead of learning a regression model to fit the spinal shape, we simply adopt the coordinates of the spine in the training samples to construct a data dictionary and formulate this problem as an $\ell_1$ norm optimization to learn the best sparse representation. Based on the regularity of the spine shape, ambiguous coordinates are removed and the sparse representation is optimized in a subspace instead of all coordinates (refer to section 2.2). Finally, the refined coordinates in each axis are reconstructed from the same subspace jointly, which further improves the localization and identification performance.

The rest of the paper is organized as follows: In section II, we introduce our deep image-to-image network architecture with message passing and refinement approach. In section III, the proposed framework is compared to previous state-of-the-art methods based on a public spine dataset. In section IV, we present the conclusion and discussion.

\section{Methodology}
\subsection{Deep Image-to-Image Network (DI2IN) for Multiple Landmark Localization}
\begin{figure}[!htbp]
    \centering
	\vspace{-4mm}
    \centerline{\includegraphics[width=\columnwidth]{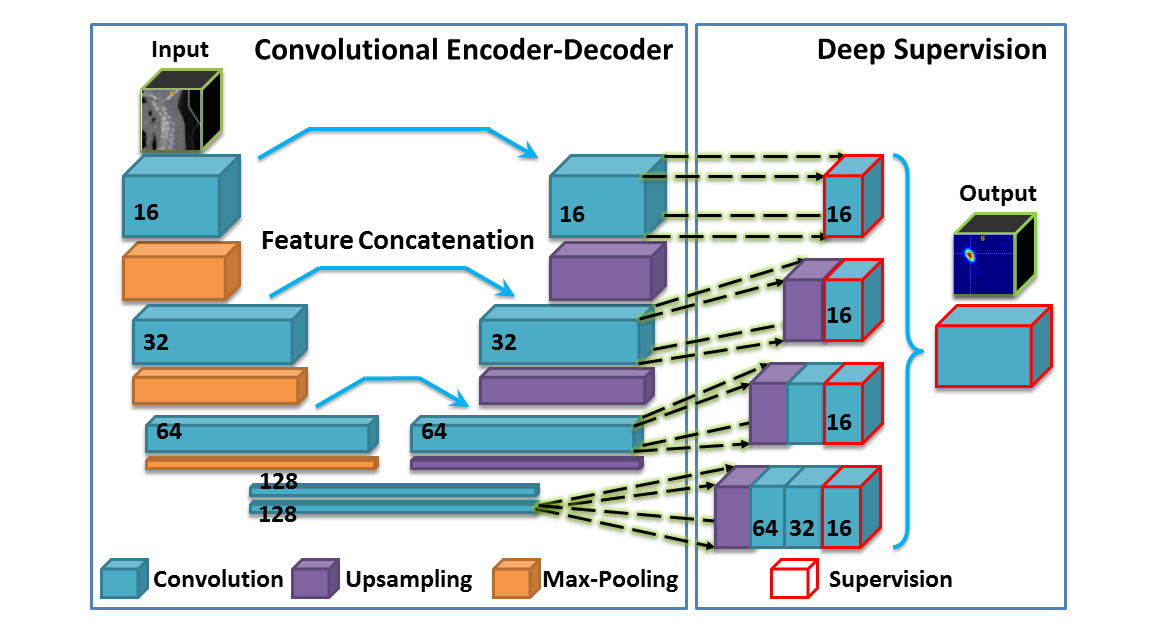}}
	\vspace{-4mm}
    \caption{Proposed deep image-to-image network (DI2IN). The front part is a convolutional encoder-decoder network with feature concatenation, and the backend is deep supervision network through multi-level. Numbers next to convolutional layers are the channel numbers.}
    \vspace{-4mm}
	\label{fig:2}
\end{figure}
In this section, we present the proposed deep image-to-image network, which is multi-layer convolutional, to localize vertebra centroids.
As shown in Figure \ref{fig:2}, the proposed network is deployed in a symmetric manner which can be treated equivalently as a convolutional encoder-decoder network.
It is implemented in the fashion of voxel-wise end-to-end learning to enable efficient inference.
The multichannel ground truth data is specially designed with the coordinates of vertebra centroid.
A Gaussian distribution $I_{gt} = \frac{1}{\sigma \sqrt{2\pi}}e^{-\left \| x-\mu  \right \|^{2}/{2\sigma^{2}}}$ is defined in each channel to represent the vertebra location, .
Vector $x\in \mathbb{R}^{3}$ represents the voxel coordinates in volume, vector $\mu$ is the ground truth location of verterbra centroid.
Variance $\sigma^{2}$ is pre-defined which controls the scale of the Gaussian distribution.
Each channel's prediction $I_{prediction}$ corresponds to a unique vertebra centroid. It has the same size as the input image.
Therefore, the whole learning problem is formulated as multichannel voxel-wise regression.
During the training, we apply the square loss $\left | I_{prediction} - I_{gt} \right |^{2}$ for each voxel at the output layer.
We define the centroid detection as a regression task instead of classification.
Because the highly imbalanced data in classification is inevitable and it causes the misleading classification accuracy.

Convolution, rectified linear unit (ReLU), and max-pooling layers are used in the encoder part of the proposed network.
Pooling is critical as it helps increase the receptive field of neurons and lower the GPU memory consumption.
With the larger receptive field, more contextual information is taken into consideration for each neuron in different layers. Therefore, the relative spatial position of vertebra centroids in prediction would be better interpreted.
The decoder part is composed of the convolution, ReLU and upsampling layers.
Upsampling layers are implemented with the bilinear interpolation to enlarge and densify the activation. It further enables the end-to-end voxel-wise training.
The convolutional filter size is $1\times1\times1$ in the final output layer and $3\times3\times3$ for the other convolution layers.
The max-pooling filter size is $2\times2\times2$.
The stride in the convolution layers is set as 1 to maintain the same size in each channel. The pooling factor in pooling layers is set as 2 for downsampling by half in each dimension.
The number of channels in each layers are marked next to the layers in Figure \ref{fig:2}.
In upsampling layers, the input features are upsampled by a factor of 2 in $x,y,z$ directions respectively.
The network takes a 3D CT image (volume) as input and directly outputs multiple probability maps, with each map associated with one vertebra landmark (equivalent to vertebra centroid).
The framework is more efficient at computing the probability maps as well as the centroid locations than the patch-wise classification or regression methods in \cite{chen2015automatic,suzani2015fast}.

Our DI2IN adopts several prevailing techniques\cite{badrinarayanan2015segnet,ronneberger2015u,xie2015holistically,merkow2016dense, dou20163d} with necessary modification.
We utilize the feature layer concatenation in DI2IN which is analogous with the one described in \cite{ronneberger2015u}.
The shortcut bridges are built up directly from the encoder layers to decoder layers.
It passes forward the feature maps from the encoder and is then concatenated with the decoder feature layers.
The concatenated features are used as the input for next convolution layers.
Following the concatenation, high and low level features are combined explicitly so that the network benefits from both the local and global contextual information.
Deep supervision in neural network during the end-to-end training is shown in \cite{xie2015holistically,merkow2016dense, dou20163d} to achieve excellent boundary detection and segmentation results.
In the network, we introduce a more sophisticated deep supervision method to improve the performance.
Several branches are bifurcated out from the main network from the intermediate layers of the decoder part.
With proper upsampling factors and convolution operations, the output size of each channel of all branches matches the size of the input image.
The supervision is introduced at the end of each branch $i$ by computing a loss term $l_{i}$ with the same ground truth data.
To further leverage the results from different branches, the final output is determined by the convolution operation of output concatenation of all branches with ReLU.
The total loss $l_{total}$ is a combination of loss terms from all output layers which includes the output layers from all branches and the final output layer, as shown here:
\[l_{total} = \sum_{i}l_{i} + l_{final}\]
\subsection{Probability Map Enhancement with Message Passing Scheme}
Given the image $I$, the DI2IN generates one probability map $P\left ( v_i | I \right )$ for the center of each individual vertebra $i$ with high confidence.
The vertebrae will be located at the peak positions $v_i$ of probability maps.
However, we find that these probability maps are not perfect yet: some probability maps don't have response or have very low response at the ground truth locations because of similar image appearances of several vertebrae (e.g. $T1\sim T12$).
In order to handle the problem of missing response, we propose a message passing scheme to effectively enhance the probability maps by utilizing the prior knowledge of the spine structure.

\begin{figure}[t]
	\vspace{-4mm}
    \centerline{\includegraphics[width=0.9\columnwidth]{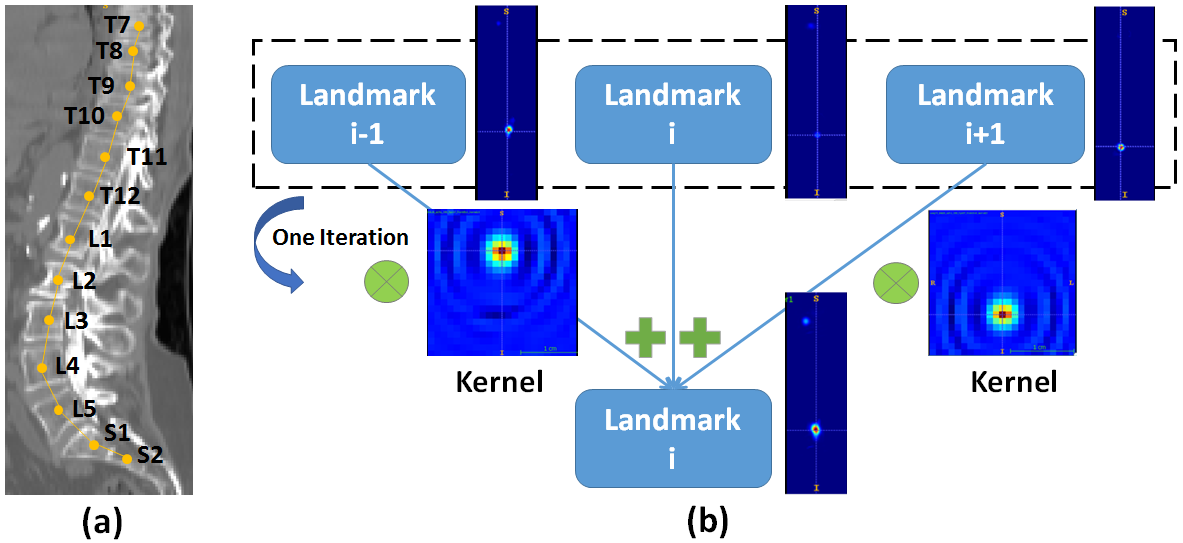}}
	\vspace{-4mm}
	\caption{(a) Chain-structure model for vertebra centers; (b) One iteration of message passing (landmarks represents vertebra centers): the neighbors' landmark probability maps help compensate the missing response of landmark i.}
	\vspace{-4mm}
    \label{fig:message_passing}
\end{figure}

The concept of message passing was first introduced in the context of probabilistic graphical models.
It is used in the sum-product or max-product algorithms for exact inference of the marginal probabilities of nodes or the distribution mode in a tree-structured graph.
Messages are passed iteratively between neighboring nodes to exchange information and optimize the overall probability distribution.
Similarly, we introduce an MRF-like model, a chain-structure graph shown in Figure \ref{fig:message_passing}, to express the spatial relationship among vertebrae, where each node in the graph represents one vertebra center $v_i$.
Then we propose the following formulation to update the $P\left ( v_i | I \right )$ during the iteration $t$ of message passing.
\begin{align}
P_{t+1}\left ( v_i | I \right )&=\frac{1}{Z}\left [ \alpha\cdot\frac{\sum_{j\in\partial i }m_{j\rightarrow i}}{\left |\partial i  \right |} + P_{t}\left ( v_i | I \right ) \right ]\\
&=\frac{1}{Z}\left [ \alpha\cdot\frac{\sum_{j\in\partial i }P_{t}\left ( v_j | I \right )\ast k\left ( v_i | v_j \right )}{\left |\partial i  \right |} + P_{t}\left ( v_i | I \right ) \right ]
\label{eq:message_passing}
\end{align}
where $\partial i$ is the neighbor of vertebra $i$ in the graph, $Z$ is a normalization constant, and $\alpha\in \left ( 0,1 \right )$ is a discounted factor.
The messages $m_{j\rightarrow i}$, defined as $P_{t}\left ( v_j | I \right )\ast k\left ( v_i | v_j \right )$, are passed along the chain shown in Figure \ref{fig:message_passing}.
$\ast$ is the convolution operation. $k\left ( v_i | v_j \right )$ is a single convolution kernel which is learned from the ground truth distribution of vertebra $i,j$.
Multi-dimensional convolution itself is capable to shift the mass of the probability map $P_{t}\left ( v_i | I \right )$ to its neighborhood with a fixed orientation (kernel).
If $P_{t}\left ( v_i | I \right )$ is confident at its correct location, then the message $m_{j\rightarrow i}$ would be a strong prior for $P_{t+1}\left ( v_j | I \right )$ at the correct location of the vertebra $j$.
After several iterations of message passing, the vertebra with missing response can be compensated with the aggregated messages from its neighboring vertebrae.
The underlying assumption is that majority of the vertebra probability maps are confident and well distributed around the true locations, which is guaranteed by the powerful DI2IN in our method.
The advantage of the proposed scheme is that it can be concatenated into the DI2IN for further end-to-end training (fine-tuning) when the iteration number is fixed.
The location of each vertebra centroid can simply be determined by the location of the maximum value in the corresponding probability map.

Several recent works have deployed the message-passing concept for different landmark detection tasks.
Chu et al.\cite{chu2016structured} proposed the passing scheme between the feature maps instead of landmark probability maps.
Yang et al.\cite{yang2016end} introduced a fully connected graphical model for message passing between probability maps.
The hand-crafted features were adopted in the pair-wise terms of the messages.
Payer et al.\cite{payer2016regressing} also brought up the fully connected graphical model, applying one-time passing with pixel-wise dot-product for noise cancelling.
In our proposed method, the passing is directly among the response maps along the chain-structure model.
The response maps are gradually enhanced within several passing iterations, since one passing is not enough to make necessary adjustment for probability maps.
Compared to the hand-craft features, the single convolutional kernel is eligible to generate messages between neighbors because the designed neighborhood is compact.
In our framework, the missing response is the major issue instead of the noisy output, so the dot-product operation is not applicable and may hurt the output probabilities.

\subsection{Sparse Representation for Landmark Refinement}
\label{sec:refine}

\begin{figure}[t]
    \centering
	\vspace{-4mm}
    \centerline{\includegraphics[width=0.7\columnwidth]{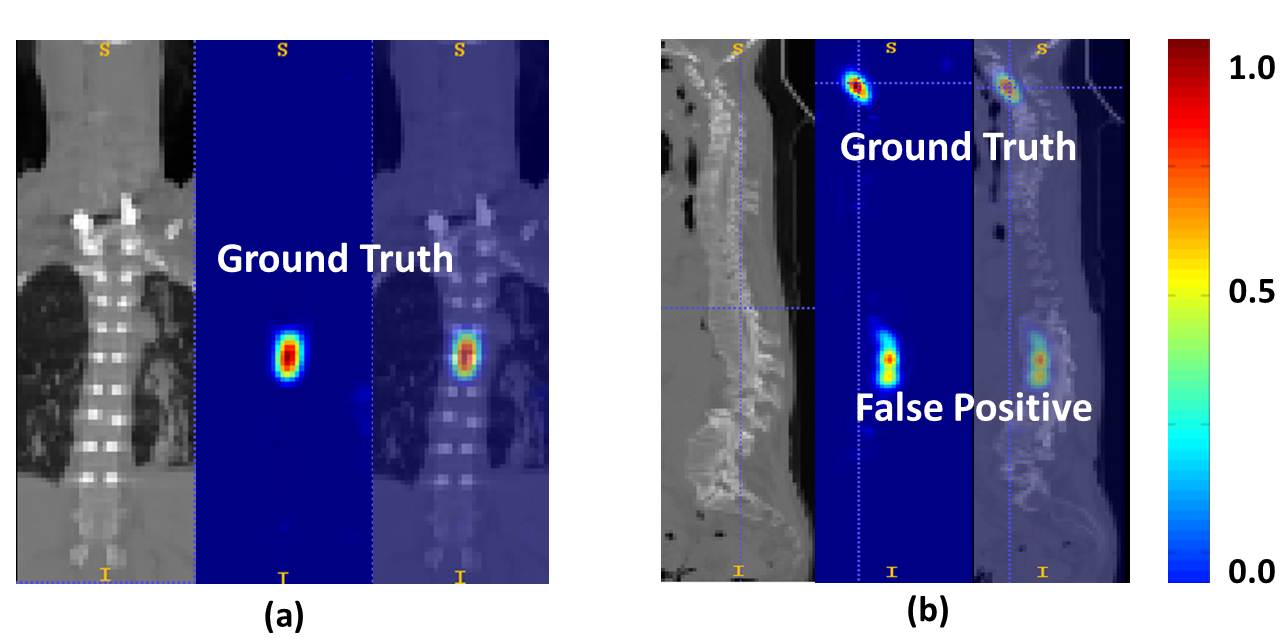}}
	\vspace{-4mm}
    \caption{Left: CT image. Middle: Output of one channel from the network. Right: Overlaid display. The prediction in (a) is close to ground truth location. In (b), a false positive response exists remotely besides the response at the correct location.}
	\vspace{-4mm}
    \label{fig:4}
\end{figure}

As shown in Figure \ref{fig:4}, the DI2IN with message passing generates a clear probability map, where the high probability map indicates the potential location of the landmark (centroid of the vertebrae). However, sometimes due to image artifacts and low image resolution, it is still difficult to guarantee there will be no false positive.
In \cite{chen2015automatic}, a shape regression optimization model was used to refine the predicted vertebral centroids in the vertical axis. By minimizing an energy function, the optimized parameters are learned for each test sample to determine the final coordinates of vertebrae. However, their model assumes that the coordinates distribution can be described in a quadratic form, and it was only applied for coordinates in the vertical axis.

Inspired by the previous works in sparse representation, we propose an $\ell_1$ norm approach to help refine the coordinates in all $x$, $y$ and $z$ axes. Given a pre-generated shape-based dictionary $\mathbf{D}$ and the predicted coordinates vector of all centroids $\mathbf{v}$ in a testing sample, we adopt the $\ell_1$ norm optimization to solve the sparse coefficient vector $\mathbf{a}$. The refined coordinates $\hat{\mathbf{v}}$ is defined as $\hat{\mathbf{v}}=\mathbf{D}\mathbf{a}$. In particular, the shape-based dictionary is learned from the training samples. For example, the dictionary $\mathbf{D}_z$ associated with the vertical axis is constructed by the $z$ coordinates of all centroids of each sample in the training database. $\mathbf{v}_z$ denotes the predicted $z$ coordinates of one sample in the testing database. The dictionaries $\mathbf{D}_x$ and $\mathbf{D}_y$ indicate the dictionaries associated with other axes and are learned in the same way.

\begin{algorithm}[ht]
\caption{The $\ell_1$ Norm Refinement}
\begin{algorithmic}[1]

\fontsize{10}{10}

\Require The dictionary $\mathbf{D}_x$, $\mathbf{D}_y$ and $\mathbf{D}_z\in\mathbb{R}^{M\times{N}}$, the predicted coordinates vector $\mathbf{v}_x$, $\mathbf{v}_y$ and $\mathbf{v}_z$ and the coefficient $\lambda$. $M$ and $N$ indicate the number of landmarks and size of items in dictionary, respectively.

\State Find the maximum descending subsequence in the predicted coordinates $\mathbf{v}_z$ by dynamic programming.

\State Add the indices associated with the maximum descending subsequence into the set $\mathcal{S}$ and define the subspace of the dictionary $\mathbf{D}_{x,\mathcal{S}}$, $\mathbf{D}_{y,\mathcal{S}}$, and $\mathbf{D}_{z,\mathcal{S}}$ and the predicted coordinates $\mathbf{v}_{x,\mathcal{S}}$, $\mathbf{v}_{y,\mathcal{S}}$ and $\mathbf{v}_{z,\mathcal{S}}$.

\State Solve the optimization problem below by $\ell_1$ norm recovery for the vertical axis $z$:

\begin{flalign*}
& \min_{\mathbf{a}_z}\frac{1}{2}||\mathbf{v}_{z,\mathcal{S}}-\mathbf{D}_{z,\mathcal{S}}\mathbf{a}_z||_2^2 + \lambda||\mathbf{a}_z||_1.
\end{flalign*}

\State Solve the same optimization problem in Step 3 for $\mathbf{v}_{x,\mathcal{S}}$ and $\mathbf{v}_{y,\mathcal{S}}$, respectively.

\State Return the refined coordinates $\hat{\mathbf{v}_x}=\mathbf{D}_x\mathbf{a}_x$,
$\hat{\mathbf{v}_y}=\mathbf{D}_y\mathbf{a}_y$ and $\hat{\mathbf{v}_z}=\mathbf{D}_z\mathbf{a}_z$.
\end{algorithmic}
\label{Algorithm}
\end{algorithm}

\begin{figure}[t]
    \centering
	\vspace{-4mm}
    \centerline{\includegraphics[width=0.7\columnwidth]{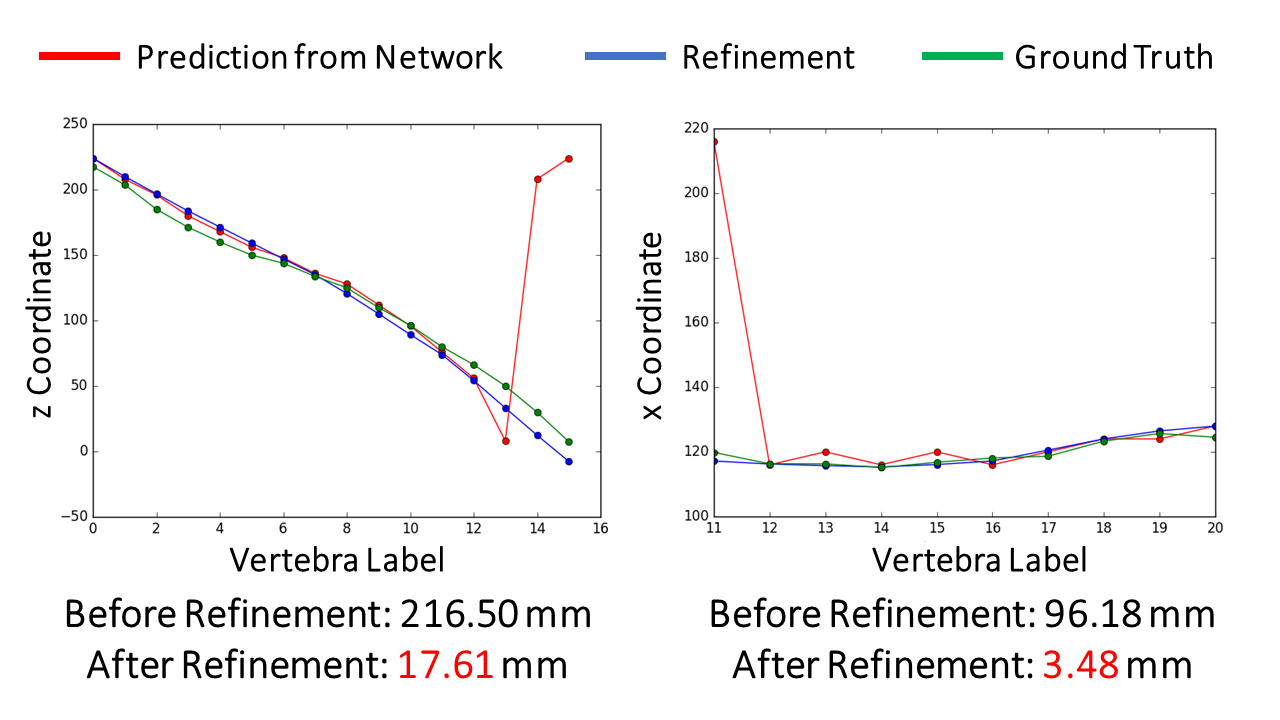}}
	\vspace{-4mm}
    \caption{Maximum errors of vertebra localization before and after the $\ell_1$ norm refinement.}
	\vspace{-4mm}
    \label{fig:3}
\end{figure}

The details are shown in Algorithm 1. First, we use dynamic programming to find the maximum descending subsequence in the predicted coordinates $\mathbf{v}_z$ since the vertical axis of the spine produces the most stable results. We define the subspace $\mathcal{S}$ of dictionary and the predicted coordinates vector based on the indices in the subsequence. For example, we only choose the atoms from dictionary $\mathbf{D}_z$ and $\mathbf{v}_z$ associated with the indices to generate a sub-dictionary $\mathbf{D}_{z,\mathcal{S}}$ and sub-vector $\mathbf{v}_{z,\mathcal{S}}$. Then we solve the optimization problem in Step 3 for $x$, $y$ and $z$ axes individually in the subspace $\mathcal{S}$ instead of the original space $\mathcal{S}_0$. Finally, all coordinates are reconstructed by the original dictionary (i.e., $\mathbf{D}_z$) and sparse vector (i.e., $\mathbf{a}_z$). Intuitively, we remove the ambiguous outliers in the preliminary predicted coordinates and then define a subspace without these outliers. Based on the subspace, we find the best sparse combination in the corresponding sub-dictionary. By taking advantage of the original dictionary, all coordinates are reconstructed and refined simultaneously as shown in Figure \ref{fig:3}.

\section{Experiments}
First, we evaluate the proposed method on the database introduced in \cite{glocker2013vertebrae} which consists of 302 CT scans of patients with varying types of pathologies.
There are several unusual appearances in the database, such as the abnormal spine curvature and the bright visual artifacts caused by metal implants from the post-operative procedures.
In addition, the field-of-view (FOV) of each CT image varies widely in terms of vertical cropping, image noise and physical resolution\cite{glocker2012automatic}.
Most cases contain a portion of whole vertebrae while the global spine structure is visible only in a few cases.
The large variations in pathologies and the limited FOV increase the complexity of vertebra appearance, and thus raise the difficulties of accurate spine localization and identification task.
The ground truth is marked at the centroid of each vertebra, which is annotated by clinical experts.
In previous works\cite{glocker2012automatic,chen2015automatic,suzani2015fast}, there are two different settings on these 302 CT images: the first one uses 112 of the images as training and another 112 images as testing; the second one takes all images (242) in setting one with extra 18 images as training data and an additional 60 images as testing data. For a fair comparison, we follow the same database settings in our experiments. They are denoted as ``Set 1'' and ``Set 2'' respectively.
We follow the evaluation metrics described in \cite{glocker2013vertebrae}, in terms of the Euclidean distance error (in mm) and identification rates (Id.Rates) defined in \cite{glocker2012automatic}.
Table 1 compares our evaluation performance with the number reported by previous approaches\cite{glocker2013vertebrae,chen2015automatic,suzani2015fast}.
We obtain an overall average mean error of 9.1 mm and 8.6 mm and an identification rates of 80\% and 85\% on those two sets, respectively.
Overall, our method outperforms the state-of-the-art methods on the same datasets in terms of mean error and identification rates.

It is well known that deep neural networks have the capability to represent the variations of a large amount of data. With large amounts of annotated data in the training, the deep neural network can usually achieve better performance on various tasks. In order to validate if more training data can boost the performance of the proposed method, we introduce additional 1000+ CT scans of patients into the training samples and train our proposed model again from scratch. These data cover large variations in populations and contrast phases which are collected for various purposes. Most cases have a large FOV and include all the vertebrae. Some scans are extended to the knee and head.
The testing data is not changed in all experiments. This pipeline is denoted as ``Our Method+1000 training data''. As shown in Table 1, the experimental results demonstrate that the large amount of training samples can further improve the performance significantly. Our approach has achieved the best performance in almost all the metrics. On ``Set 1'', the Id. Rates of our method is 13 percent higher than the state-of-the-art method\cite{glocker2013vertebrae}. We also achieve more than 90\% Id. Rates on ``Set 2'', which is 6 percent higher than the state-of-the-art method\cite{chen2015automatic}.

\begin{table}[!htbp]
\renewcommand{\arraystretch}{1.3}
\renewcommand{\multirowsetup}{\centering}
\setlength{\belowrulesep}{0pt}
\setlength{\aboverulesep}{0pt}
\caption{Comparison of localization errors in $mm$ and identification rates among different methods. ``Set 1'' has 112 CT images for training and 112 images for testing. ``Set 2'' uses all data in ``Set 1'' with extra 18 images for training and 60 images for testing. Our Method (DI2IN+MP+Sparsity) is trained and tested using default data setting in ``Set 1'' and  ``Set 2'', while ``+1000'' indicates this model is trained with additional 1000 images and evaluated on the same testing data. Evaluation of results after each step are also listed for comparison, which shows that they improve the performance. ``MP'' and ``Sparsity'' denote message passing scheme and sparsity regularization respectively.}
\begin{center}
\begin{tabular}{|c|c|c|c|c|c|c|c|}
\hline
\multirow{2}{*}{Region} & \multirow{2}{*}{Method} & \multicolumn{3}{|c|}{Set 1} & \multicolumn{3}{|c|}{Set 2}\\
\cmidrule{3-8}
& & Mean & Std & Id.Rates & Mean & Std & Id.Rates\\
\cmidrule{1-8}
\multirow{9}{*}{All} & Glocker \emph{et al.}\cite{glocker2013vertebrae} & 12.4 & 11.2 & 70\% & 13.2 & 17.8 & 74\% \\
\cmidrule{2-8}
& Suzani \emph{et al}\cite{suzani2015fast} & 18.2 & 11.4 & - & - & - & - \\
\cmidrule{2-8}
& Chen \emph{et al.}\cite{chen2015automatic} & - & - & - & 8.8 & 13.0 & 84\%\\
\cmidrule{2-8}
& DI2IN & 17.0  & 47.3 & 74\%& 13.6 & 37.5  & 76\%  \\
\cmidrule{2-8}
& DI2IN+MP & 11.7 & 19.7 & 77\%& 10.2 & 13.9 & 78\%  \\
\cmidrule{2-8}
& DI2IN+MP+Sparsity & 9.1 & 7.2 & 80\%& 8.6 & 7.8 & 85\%  \\
\cmidrule{2-8}
& DI2IN+1000 & 10.6 & 21.5 & 80\%& 7.1 & 11.8 & 87\%  \\
\cmidrule{2-8}
& DI2IN+MP+1000 & 9.4 & 16.2 & 82\%& 6.9 & 8.3 & 89\%  \\
\cmidrule{2-8}
& DI2IN+MP+Sparsity+1000 & \bf{8.5} & \bf{7.7} & \bf{83}\% & \bf{6.4} & \bf{5.9} &\bf{90\%} \\
\hline
\multirow{5}{*}{Cervical} & Glocker \emph{et al.}\cite{glocker2013vertebrae} & 7.0 & 4.7 & 80\% & 6.8 & 10.0 & 89\%\\
\cmidrule{2-8}
& Suzani \emph{et al}\cite{suzani2015fast} & 17.1 & 8.7 & - & - & - & - \\
\cmidrule{2-8}
& Chen \emph{et al.}\cite{chen2015automatic} & - & - & - & \bf{5.1}  & 8.2 & 92\%\\
\cmidrule{2-8}
& DI2IN+MP+Sparsity & 6.6 & 3.9 & 83\% & 5.6 & \bf{4.0} & 92\% \\
\cmidrule{2-8}
& DI2IN+MP+Sparsity+1000 & \bf{5.8} & \bf{3.9} & \bf{88}\% & 5.2 & 4.4 &\bf{93\%}\\
\hline

\multirow{5}{*}{Thoracic} & Glocker \emph{et al.}\cite{glocker2013vertebrae}  & 13.8 & 11.8 & 62\% & 17.4 & 22.3 & 62\%\\
\cmidrule{2-8}
& Suzani \emph{et al}\cite{suzani2015fast} & 17.2 & 11.8 & - & - & - & -\\
\cmidrule{2-8}
& Chen \emph{et al.}\cite{chen2015automatic}  & - & - & -  & 11.4 & 16.5  & 76\%\\
\cmidrule{2-8}
& DI2IN+MP+Sparsity & 9.9 & 7.5 &  74\%& 9.2 & 7.9 & 81\% \\
\cmidrule{2-8}
& DI2IN+MP+Sparsity+1000 & \bf{9.5} & \bf{8.5} & \bf{78}\% & \bf{6.7} & \bf{6.2} &\bf{88\%}\\
\hline

\multirow{5}{*}{Lumbar} & Glocker \emph{et al.}\cite{glocker2013vertebrae}  & 14.3 & 12.3 & 75\%  & 13.0 & 12.5 & 80\%\\
\cmidrule{2-8}
& Suzani \emph{et al}\cite{suzani2015fast} & 20.3 & 12.2 & - & - & - & - \\
\cmidrule{2-8}
& Chen \emph{et al.}\cite{chen2015automatic}  & - & - & -  & 8.4 & 8.6  & 88\%\\
\cmidrule{2-8}
& DI2IN+MP+Sparsity & 10.9 & 9.1 & 80\%& 11.0 & 10.8 & 83\%\\
\cmidrule{2-8}
& DI2IN+MP+Sparsity+1000 & \bf{9.9} & \bf{9.1} & \bf{84}\% & \bf{7.1} & \bf{7.3} &\bf{90\%} \\
\hline

\end{tabular}
\end{center}
\end{table}

All experiments are conducted on a workstation equipped with an Intel 3.50 GHz CPU and a 12GB Nvidia Titan X GPU.
During the evaluation, the response maps of all output channels are compared with a heuristic threshold constant in an element-wise manner in order to distinguish valid response from random noise.
Only the channels whose response maps contain elements with value greater than the threshold are considered.
The vertebra centroids associated with these channels are then identified to be present in the image. The landmarks corresponding to the other response maps are considered as non-presented. The localization and identification of all vertebrae in one case is achieved simultaneously in an efficient way.
The testing time of our method is around three seconds per case on average assisted with the GPU.
The experimental results demonstrate that our proposed method for spine centroids localization and identification is not only effective in terms of accuracy, but also significantly time-efficient.

\section{Conclusion}
In this paper, we proposed an effective and fast automatic method to localize and label vertebra centroids in 3D CT volumes.
Our method outperforms other state-of-the-art methods of spine labeling in terms of various evaluation metrics.
For the future study,  we plan to investigate various DI2IN architectures (e.g. ResNet) and other sophisticated refinement approaches to further improve the localization and identification performance.

\vspace{2mm}
\noindent\textbf{Disclaimer}: This feature is based on research, and is not commercially available. Due to regulatory reasons its future availability cannot be guaranteed.
\end{document}